# Proceedings of the Workshop on Social Robots in Therapy: Focusing on Autonomy and Ethical Challenges

**Abstract**

Robot-Assisted Therapy (RAT) has successfully been used in HRI research by including social robots in health-care interventions by virtue of their ability to engage human users both social and emotional dimensions. Research projects on this topic exist all over the globe in the USA, Europe, and Asia. All of these projects have the overall ambitious goal to increase the well-being of a vulnerable population. Typical work in RAT is performed using remote controlled robots; a technique called Wizard-of-Oz (WoZ). The robot is usually controlled, unbeknownst to the patient, by a human operator. However, WoZ has been demonstrated to not be a sustainable technique in the long-term. Providing the robots with autonomy (while remaining under the supervision of the therapist) has the potential to lighten the therapists burden, not only in the therapeutic session itself but also in longer-term diagnostic tasks. Therefore, there is a need for exploring several degrees of autonomy in social robots used in therapy. Increasing the autonomy of robots might also bring about a new set of challenges. In particular, there will be a need to answer new ethical questions regarding the use of robots with a vulnerable population, as well as a need to ensure ethically-compliant robot behaviours. Therefore, in this workshop we want to gather findings and explore which degree of autonomy might help to improve health-care interventions and how we can overcome the ethical challenges inherent to it.

**Index Terms**

Autonomous Robots; Robots in Therapy; Ethics; Human-Robot Interaction.

## I. Introduction

It has been proposed that robots in future therapeutic scenarios should be capable of operating autonomously (sometimes while remaining under the supervision of the therapist) for at least some of the time. A fully autonomous robot might be able to infer and interpret a patients intentions in order to understand their behavior and provide real-time adaptive behavior given that patients individual needs. However, full autonomy (in the sense that the robot can adapt to any event during the therapeutic sessions) is currently unrealistic and not desired as the robots action policy will not be perfect and in certain therapeutic scenarios, every single action executed by the robot should be appropriate to the therapeutic goals, context of the interaction, and the state of the patient. It is the aim of this workshop to reflect on existing RAT robots and ongoing research on HRI in the domestic and care facility contexts. A lot of work has already been completed to understand user needs and design more autonomous robot behaviors, as well as to build platforms and evaluate them in terms of acceptance and usability. In this workshop we want to gather, compare, and combine knowledge gained in various HRI projects with robots in therapy in the US, Europe, and Asia. This should lead to a broader understanding of how increasing the degree of autonomy of the robots might affect therapies as well as the design and ethical challenges of health-care robots

## II. Invited Speakers

- **Adriana Tapus**, Robotics and Computer Vision Lab, ENSTA-ParisTech
- **Maartje de Graaf**, Humanity Centered Robotics Initiative, Brown University
- **Jenay Beer**, Institute of Gerontology, University of Georgia
- **Jamy Li**, Human Media Interaction, University of Twente

## III. Organizing committee

- **Pablo Gómez Esteban**, Robotics and Multibody research group, Vrije Universiteit Brussel, Belgium
- **Daniel Hernández García**, Centre for Robotics and Neural Systems, University of Plymouth, UK
- **Hee Rin Lee**, Computer Science and Engineering, UC San Diego, USA
- **Pauline Chevalier**, Human Media Interaction, University of Twente, The Netherlands
- **Paul Baxter**, Lincoln Centre for Autonomous Systems, University of Lincoln, UK
- **Cindy Bethel**, Computer Science and Engineering Department, Mississippi State University, USA

# Emotional State Recognition for Personalized Robot Behavior Adaptation: Ethical Implications

Jainendra Shukla*
Instituto de Robótica para la Dependencia
Barcelona

Joan Oliver
Instituto de Robótica para la Dependencia
Barcelona

Domènec Puig
Intelligent Robotics and Computer Vision Group (IRCV), Rovira i Virgili University
Tarragona

## ABSTRACT
Recent advancements in the Socially Assistive Robotics (SAR) have shown a vital potential and have thus inspired us to explore the benefits of robot-assisted cognitive stimulation. In this paper, we have argued that SAR based cognitive stimulation interactions work mainly on the cognitive level of the target user. Therefore, ethical implications around such robot interactions goes beyond the physical level and hence, require a new set of regulations and guidelines.

## KEYWORDS
Adaptation; rehabilitation; cognitive stimulation; intellectual disability

## 1 INTRODUCTION

Socially Assistive Robotics (SAR) has already been widely used in mental health service and research, primarily among children with Autism Spectrum Disorder (ASD) and among older adults with dementia. Motivated by the benefits offered by SAR in mental health service & research, we envision that SAR can also benifit cognitive rehabilitation of individuals struggling with a wide range of mental health concerns. Among cognitive rehabilitation approaches used, Cognitive Stimulation Therapy (CST) is an evidence-based psychological or psychosocial intervention consisting of structured sessions of stimulating activities in a group setting or for individuals [6]. CST can serve users with different mental conditions, including older adults with dementia, adults with intellectual disability (ID) etc. Consequently, SAR empowered CST can positively affect well-being of wide variety of users.

It has been shown that a robot with adaptive behavior can improve the user's task performance in the cognitive activities [7]. Moreover, the robot behavior adaptation can be based on the patient's level of disability and their current level of emotional state and engagement.

## 2 EMOTIONAL ADAPTIVE BEHAVIOR

We have focused on robot behavior adaptation based upon the current level of emotional state and engagement. Accordingly, we proposed an interactive and adaptive architecture of SAR assisted mental health interventions (fig. 1). The framework in figure 1 can be understood as follows:

*jshukla@institutorobotica.org

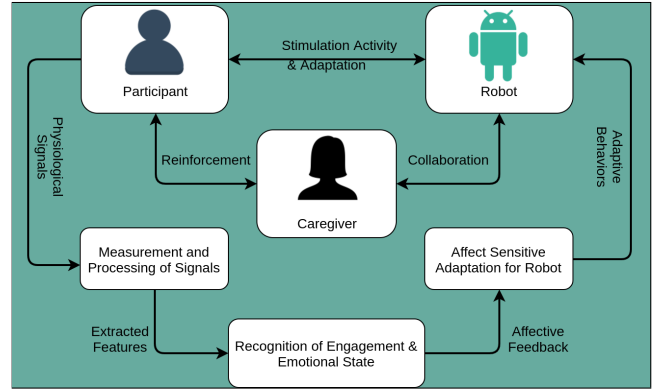

Figure 1: SAR empowered cognitive stimulation.

(1) Based upon the user's requirements and needs firstly the caregiver identifies a specific stimulation activity. Consequently, the robot collaborates with the caregiver during the execution of the designed activity with the participant.
(2) Participants are equipped with different physiological sensors to obtain biosignals such as Heart Rate (HR), Electrodermal Activity (EDA) etc. High resolution video cameras can also be employed to obtain the audiovisual recordings to interpret vocal and facial expressions.
(3) Various signal processing techniques are applied online to extract meaningful features from the above signals and camera data.
(4) Automatic emotional state classification and engagement prediction is done from above features using assorted machine learning and data mining algorithms.
(5) The above information is fed to the robot, based upon which robot plans a new course of action that can be either to continue the activity or to encourage the participant or to halt the activity.
(6) The robot takes the newly planned actions based upon the feedback obtained from the participant in collaboration and agreement with the caregiver. The specificity of the communication between the robot and the caregiver can be determined based upon the technical abilities of the robot.

Wearable sensors must be hassle free to obtain the physiological signals in real-time for the participants and should cause a minimum level of distraction, as the targeted users are individuals with ID, who can be easily distracted by any device or sensor attached to their body or placed around them. Any distraction caused by these sensors will lead to adverse effects on the productivity of interaction activities. Sample wearable devices that can be used fitting these restrictions are E4 wristband[1] for recording EDA and HR data, Emotiv Epoc+[2] for recording Electroencephalogram(EEG) data etc. Now, the research is in the second phase, which is *learning & adaptation of engagement/emotional state*. It aims to provide the automated detection of engagement and emotional state ability to the robots in real time during stimulation interventions.

## 3 ETHICAL IMPLICATIONS

While most Europeans have a positive impression about the robots but there has been a significant decline in the proportion of the respondents holding the positive view since 2012, from 70% down to 64% [5]. At the same time, a large proportion of the survey respondents still were uncertain about the potential social or medical uses of the robots [5]. In contrast, during the project, we encountered different trend about the use of SAR for cognitive stimulation upon interviewing at-least 7 experts. These experts are professional caregivers working with individuals with wide range of mental health concerns. Not only the experts were mostly positive about the use of the robots but were also optimistic about the potential benefits of the robots for cognitive stimulation interventions.

The ethical, legal and social implications (ELSI) of the robots interacting at a physical level has received significant attention in the past. In particular, safety requirements on several design factors such as robot shape, robot motion, incorrect autonomous decisions etc. have been specified [1]. But the human-robot interaction (HRI) of the proposed system works not only at the physical level but also at the cognitive level. This certainly increases the complexity of the issues involved. Accordingly, several questions about ethical implications of the system arose during the ongoing course.

**How does the processing of user's emotional data (obtained via physiological signals and/or external sensors) challenges the necessity and proportionality principles?**

Mere acquisition and collection of this extremely sensitive information can challenge the proportionality and necessity principles. Necessity requires that the researchers must first seek the consent of the parents and/or the legal guardians of the users involved in the study. It is the first step before assessing the proportionality of the limitation. It inquires about the required amount of information to allow the robot to emotionally adapt [2]. During the analysis phase, the experts explained the characteristics of the individuals with ID and it was pointed out that the users have limited ability in recognition and expression of their emotional state. Hence, it was agreed upon that the analysis of physiological and behavioral correlation of emotions is the most useful method for monitoring their emotions [3]. At the same time, to limit the concerns related with the violation of user privacy and data storage, it was agreed that the emotional data will be processed off-line for the machine learning only. Later, to provide the adaptive behavior to the robot, we will capture and process the emotional data in online fashion only.

---

[1]https://www.empatica.com/e4-wristband
[2]https://www.emotiv.com/epoc/

**Who will decide about the proportionality of such information when the target users have limited intellectual abilities?**

This is another related question with the necessity & proportionality principles. Usually, the parents and/or legal guardians are responsible for decision making about their wards. But this project aims to employ clinical interventions at the cognitive level and hence several key stakeholders have been identified to discuss the issue at hand. These include robotics researchers, families of the target users, health-care providers (e.g. medical doctors, caregivers, care facilities etc.) and professional organizations (e.g. social security entities, regulatory agencies etc.).

**Will empowering the robots with emotional adaptive behavior motivate for replacement of the human caregivers?**

It is a basic question that has been the center of the discussion since the evolution of the robots. While it is easier to replace the humans in the industrial environments and it is for this reason that the robots have been installed successfully in these environments (e.g. automated manufacturing and assembly). But the use of SAR cannot replace caregivers at residential care facilities. However, it has been proven that it can help them to focus their time and resources on providing better monitoring and personalized attention of the individuals [4]. Caregivers need specific training to take maximum advantage of SAR in the health care [4]. It will further help to reduce fear of getting replaced by the robots.

## 4 FUTURE WORK

We have argued that SAR based cognitive stimulation interactions work mainly on the cognitive level of the target users. Hence, the ELSI around the robot goes beyond the physical level which cannot be comprehended with the current regulations and guidelines. Accordingly, a new set of guidelines are required for the cognitive interactions of SAR in therapeutic contexts.

## ACKNOWLEDGMENTS

This research work has been supported by the Industrial Doctorate program (Ref. ID.: 2014-DI-022) of AGAUR, Govt. of Catalonia.

# Towards a Computational Model of Dignity to Guide Autonomous Assistive Robots

Jason R. Wilson
Northwestern University
jrw@northwestern.edu

Linda Tickle-Degnen
Tufts University
linda.tickle_degnen@tufts.edu

## 1 INTRODUCTION

There are many challenges in designing assistive robots, and as we consider how these robots may be more autonomous, there are many safety and ethical issues that arise. Many ethical issues stem from ensuring that the people being assisted by the robot are treated properly, with respect and dignity, and that ultimately the assistance being provided leads to a better quality of life. Issues involving personal dignity – including privacy, autonomy, freedom from pain, and sense of identity – are highly pertinent in designing assistive robots. In particular, it is paramount that the autonomy of the robot not impinge upon the autonomy of the individual.

We introduce here the first steps to developing a computational model of dignity that an assistive robot may use to guide how it provides assistance. We focus here on modeling one influence of dignity, that of human autonomy, and describe how an autonomous robot may use the model to influence how it selects an appropriate assistance. We describe our work in the context of occupational therapy and a robot assisting an older adult.

## 2 BACKGROUND

Dignity has to do with how a person feels as a human being, how one is respected, and the autonomy with which one is able to live. Scholars have made a distinction between two forms: (1) *Menschenwürde*, an inviolable dignity that is inherent in all human beings and not dependent on behavior, beliefs, or circumstances, [3, 4, 7, 8] and (2) other forms of dignity that may be possessed to varying degrees [8]. *Menschenwürde* describes the basic dignity that all humans have simply by being human. It cannot be taken away, as much as one cannot make one no longer a human.

One form of dignity that may be possessed to varying degrees is *Dignity of Identity*, which has to do with the self-respect we have for our individuality. It is the dignity we have as individuals with autonomy, integrity, a history and a future, and relationships with other people [4]. Dignity of Identity can be affected by the behavior of others and can be impacted by injury, illness, or old age [3, 4]. When a person loses the ability to do tasks with competence (perhaps due to injury, illness, or old age), the person may receive assistance in these tasks. If the assistance does not match the needs of the person, the assistance may inhibit the person's ability and desire to choose how to accomplish the task (or even whether to do the task).

The loss of autonomy can impact one's sense of personal dignity. A person's autonomy can be diminished when the person is not able to do what he or she wants or is entitled to do. This restriction may be the result of external forces such as being physically constrained, or it may be the result insulting, hurting, or hindering the person [4].

While autonomy and the right to make decisions applies to all activities, it especially applies to therapeutic care. As much as possible and as long as an individual demonstrates the necessary cognitive, social, and emotional capabilities, an individual should be free to exercise his or her own judgment in how to adhere to a care plan. A person or robot assisting in the therapy should respect the choices made by an individual while still providing assistance to guide the person to follow essential therapeutic plans and ultimately improving health and quality of life.

Many health-care professionals are aware of the extra care that may be necessary to help preserve the sense of dignity, particularly with their older adult clients. Many older adults are referred to occupational therapists (OTs), who provide assistance while helping a person acquire or reacquire the skills necessary for everyday activities (e.g., bathing, dressing, managing medications). For some of these activities, a robot could provide some assistance, but the robot needs to do so in a manner that respects the person and treats the person with the same dignity that an OT would.

A practical guide used in occupational therapy for assessing how much assistance to provide is the levels of assistance described in the Performance Assessment of Self-care Skills (PASS) manual [5]. It defines 9 levels of assistance, ranging from verbal support at level 1 to complete assistance at level 9. The guide describes an approach in which assistance is provided only when needed to progress a task forward [5]. An advantage of this approach is that minimal assistance may maximize the autonomy of the individual. One way to maximize the autonomy of the individual is to minimize the amount of assistance provided. As assistance increases, the individual may grow dependent on the assistance which can reduce how much the individual is self-reliant, i.e., how much the individual uses his or her own capabilities (physical, cognitive, or otherwise) to execute a task. This dependency restricts or constrains the individual, thus limiting his or her autonomy. Conversely, minimizing the assistance requires a person to rely on his or her own capabilities and allows the individual to freely decide and act. However, too little assistance can also have a negative effect on autonomy. Thus, is it important to match the level of assistance to the capabilities of the individual to maximize autonomy.

Having a robot provide just the right amount of assistance and enabling the person to do tasks on her own are important elements to supporting and maintaining autonomy of the individual. Examples of robots using incremental or graded assistance include social robots to aid in medication sorting [9], ASD therapy [1], hand washing for a person with dementia [2], and physical therapy for post-stroke rehabilitation [6]. An end goal in each of these works is to only provide just enough assistance and not immediately jump to the most complete or direct assistance, thereby allowing the individual to figure things out for him or herself and thus supporting the autonomy of the person.

## 3 COMPUTATIONAL MODEL

An architecture for controlling an autonomous assistive robot needs to have a component that determines how the robot behaves. To ensure the dignity of the person being assisted is preserved, this component needs to be able to monitor the person and take actions that ensure the dignity of the person is preserved.

One aspect of dignity the robot can consider is how its actions influence the autonomy of the person being assisted. If the robot minimizes how much assistance it provides, then the person must attempt to achieve the task on his or her own and thereby maximizing the person's autonomy. However, too little assistance can also have a negative effect, such as leaving a person not knowing what to do. Thus, the robot needs to provide an optimal amount of assistance.

Consider a scenario in which a robot is assisting a person with a medication management task involving organizing medications into a sorting grid (see [9]). When a person misplaces a pill (places it in a wrong position in the grid), the robot needs to decide if and how to assist. If a robot gives no indication of a mistake, this could allow the person to notice the mistake without any assistance. Conversely, if the robot does indicate there is a mistake, perhaps saying, "I think you made a mistake somewhere," but the person does not know where the mistake is or how to fix it, then the person could feel lost or hopeless, begin to lose confidence, and possibly lose the desire to correctly complete the task. Ideally, the robot identifies how much difficulty the person is having and provides an assistance that matches that need.

We define optimal amount of assistance as the assistance that minimizes the difference between the need the person has and the assistance the robot provides. We assume here that measures of need and assistance can be normalized to be on the same scale.

For a robot to estimate the need of a person, it may consider how much progress a person is making towards a goal, how long it has been since the person made progress, and whether the person signals that there is a need. Signals may include the person explicitly requesting help, or it may be more subtle like looking at the robot, perhaps with a confused look. Also, the need a person has is often related to how much need the person recently had. For example, as a person continues to not make progress, their need incrementally increases beyond their previous need.

The robot also needs to be able to estimate how much assistance it is providing. For this we turn to the levels of assistance as defined in the PASS Manual [5]. A minimal amount of assistance would be verbal support, such as the robot saying, "Yes" or "Keep going." Each action that the robot may take to assist a person must be assigned a level of assistance.

In deciding which action to take, given a set of actions that provide correct assistance, the robot selects the action $a$ based on the equation $\arg\min_a f(a) = need(p) - assistance(a)$, where $need(p)$ is the need of the person being assisted and $assistance(a)$ is the amount of assistance provided by action $a$. For example, in the scenario of a person misplacing a pill, if this is the first mistake, then the robot may infer a low amount of need (i.e., 2) and correspondingly provide a level 2 assistance (i.e., "I think you made a mistake."). If the person hesitates, looks at the robot for more guidance, then the robot may infer the person has more need (i.e., 3) and give a level 3 assistance (i.e., "Try moving the blue pill on Tuesday.").

If the need of person perfectly matches the assistance provided, then the actions of the robot should not affect the person's autonomy. To monitor the impact of the robot's actions on the person's autonomy, we estimate autonomy as follows:

$$\textbf{autonomy}(p) = \frac{c - |need(p) - assistance(a)|}{c}$$

where $c$ is constant that represents the maximum distance between the estimated need and the level of assistance the robot provides. Ideally, autonomy remains close to 1.0, indicating that dignity in respect to autonomy is being preserved.

## 4 CONCLUSION

We presented here the first steps towards a computational model of dignity that an autonomous, assistive robot may use to guides how it assists in a therapeutic environment. So far we have focused on how the autonomy of the person being assisted influences a person's dignity. To maximize the autonomy of the person, we proposed that a robot selects actions such that the assistance provided matches the need of the person.

# Towards a Full Spectrum Diagnosis of Autistic Behaviours using Human Robot Interactions

Madeleine Bartlett
University of Plymouth
Plymouth PL4 8AA
madeleine.bartlett@plymouth.ac.uk

Tony Belpaeme
University of Plymouth
Plymouth PL4 8AA
tony.belpaeme@plymouth.ac.uk

Serge Thill
University of Plymouth
Plymouth PL4 8AA
serge.thill@plymouth.ac.uk

## ABSTRACT
Autism Spectrum Disorder (ASD) is conceptualised by the Diagnostic and Statistical Manual of Mental Disorders (DSM-V) [1] as a spectrum, and diagnosis involves scoring behaviours in terms of a severity scale. Whilst the application of automated systems and socially interactive robots to ASD diagnosis would increase objectivity and standardisation, most of the existing systems classify behaviours in a binary fashion (ASD vs. non-ASD). To be useful in interventions, and to overcome ethical concerns regarding overly simplified diagnostic measures, a robot therefore needs to be able to classify target behaviours along a continuum, rather than in discrete groups. Here we discuss an approach toward this goal which has the potential to identify the full spectrum of observable ASD traits.

## 1 INTRODUCTION
Autism Spectrum Disorder (ASD) is defined by the DMS-V in terms of two behavioural domains: social communication and interaction, and restricted or repetitive behaviours and interests [1]. Recent advances in our understanding have led to the re-conceptualisation of ASD as a spectrum. This concept refers to: (1) differences in presentation and severity within the clinical population, (2) the continuous distribution of "autistic traits" between the general and clinical populations, and (3) subgroups [6]. Diagnosis of ASD cannot, therefore, be thought of as a binary classification (e.g. non-ASD vs. ASD) but rather in terms of severity scales applied to multiple behaviours and traits. Diagnosis thus relies largely on subjective interpretations of various sources of information [2, 10], and children with ASD demonstrate high levels of clinical heterogeneity [4, 11]. The diagnostic standard of ASD could, therefore, be improved by more quantitative, objective measures of social response.

These benefits can be provided by introducing automated systems into the diagnostic process in the form of socially interactive robots [3], and systems to aid in the diagnosis of several behavioural and psychological disorders including ASD [7, 12] have been developed. However, in contrast with the diagnostic requirements, these systems usually approach behaviour classification in a binary fashion; individuals are classed as either *ASD* or *non-ASD* [12]. This lack of sensitivity to intermediate cases brings with it the ethical issues of overly simplified diagnostic measures, such as potentially classifying a large proportion of the behaviours which fall on the autism spectrum as non-ASD [7]. Here, we discuss an approach toward, and the benefits of, non-binary, automated classification of autistic behaviours embedded within human-robot interactions.

## 2 ROBOTS AS DIAGNOSTIC TOOLS FOR ASD
The prospect of introducing robots into interventions for ASD has become increasingly popular due to findings indicating that robots can promote motivation, engagement, and the occurrence of otherwise rare social behaviours in children with ASD [2, 14]. They have therefore been proposed as an effective tool for helping children develop and employ social skills, and to transfer these skills to interactions with humans [2, 13]. Whilst less attention has been given to the role of robots in ASD diagnosis [14], such an application of robot technology does offer unique benefits including: (1) standardisation of stimulus and recording methodology, and (2) increased repeatability [2, 8]. It has also been argued that a robot's ability to generate social prompts allows for the controlled elicitation and examination of social responses [2]. This is in-line with the goal of diagnostic instruments such as the Autism Diagnostic Observation Schedule (ADOS) [5], i.e. to elicit spontaneous behaviours in a standardised context. Furthermore, the finding that children with ASD interact more with technology than with humans [8] indicates that having a child interact with a robot during assessment may facilitate the production of a wider range of behaviours. This facilitation could, in turn, provide richer data for the purposes of diagnostic analysis [14].

On-line behaviour adaptation is important for autonomous robots in ASD interventions due to the high variability seen between children with ASD [3]. This process requires the system to track and classify the child's behaviour before appropriate responses can be selected. However, many systems which are used to classify behaviours in therapeutic settings are limited to simple, easily distinguished classes; they do not identify intermediate classes [12]. Wall and colleagues [12] used a subset (8 out of 29) of behaviours coded from ADOS to design a diagnostic algorithm which could differentiate between children with and without ASD. Whilst the algorithm could classify cases correctly, Wall and colleagues simplified the problem by removing the middle diagnostic classes, leaving only ASD and non-ASD. As a result, individuals who fall in the middle of the ASD spectrum were identified as non-ASD. Furthermore, an attempt to replicate these findings found that the algorithm was not robust enough to deal with a different dataset and a larger group of coded behaviours was required to identify individuals diagnosed as being in a mid-spectrum ASD class [7].

The spectrum nature of ASD means that to avoid underidentification and to allow the system to provide useful information for decisions about therapeutic approaches, classes of behaviour which do not fall at the extremes of the spectrum, e.g. High-Functioning Autism, should be identifiable. Contemporary approaches to non-binary classification are rare. Bone and colleagues [7] used a similar machine learning method to that of [12], but incorporated all the behaviour codes from ADOS which made the classification system more robust and more accurate. Including the middle diagnostic classes did decrease the accuracy but it still remained high (i.e. 96% dropped to 82%). However, this approach is still labor intensive and time-consuming, and is designed to be run off-line using data collected by the clinician.

## 3 CLASSIFYING CONTINUOUS BEHAVIOURS USING CONCEPTORS

For a classification system to accurately identify the intermediate classes of ASD, it must be able to classify behavioural patterns ranging from "typical of the general population" to "severely atypical". This can be achieved using purely machine learning methods. However, this requires a large, representative data-set which is often difficult and time-consuming to obtain due to the need to annotate the training data-sets. We therefore require a methodology that can deal with the spectrum nature of ASD by representing behaviours over continuous dimensions, and which requires less data for learning than traditional machine learning methods. One approach is to use conceptors [9]; neuro-computational mechanisms that can be used for learning a large number of dynamical patterns. Conceptors can also be combined and morphed to generate new patterns based on learned prototypical extremes along a behavioural continuum, e.g. a system given the prototypes for "walking" and "running" can generate patterns for "jogging" [9]. This approach assumes that there is a continuum underlying the behaviour, which is well suited to the symptomology of ASD [1], as demonstrated by ADOS [5] which scores behaviours such as speech abnormalities on a scale of 0 ("no evidence of abnormality") to 3 ("markedly abnormal").

To represent the spectrum nature of ASD using conceptors, a recurrent neural network can be provided, for example, with the prototype patterns for typical and markedly abnormal speech behaviour. Relevant information from these input patterns are then represented as the internal state of the system. These internal states are then used for classification, rather than the inputs themselves. Conceptors can be computed to represent the state of each dimension of speech (volume, intonation, stress, etc.) within each pattern, and clustered to form groups. These groups represent the key components of the behavioural continuum which are described by the prototype patterns provided. Morphing of these patterns using linear mixes of the prototype conceptors allows the system to interpolate less extreme patterns into the representational continuum for the behaviour. When provided with inputs of behaviours which fall in the middle of this continuum, the system already has a representation of the internal state this input would provoke, and can classify that input according to the continuum, rather than into a discrete class.

## 4 DISCUSSION AND CONCLUSIONS

In this paper we have briefly discussed how conceptors could provide an alternative to machine learning methods of automated behaviour classification for ASD diagnosis. By representing behaviours as continuous, the proposed approach has the potential to identify a more complete spectrum of ASD behaviours, rather than just extreme behaviours. Implementing such a system within a socially interactive robot would also leverage those benefits, providing a control system able to more accurately assess child behaviour to inform response selection, as the robot would be able to appropriately select and perform social prompts to elicit behaviours from the child in a standardised and repeatable manner. This application accommodates the goals of diagnostic models, e.g. ADOS [5]. Our next steps are to develop such a system, based on data from the DREAM project [1] [13], to train the system and test its performance.

## ACKNOWLEDGMENTS

This work is part of the EU FP7 project DREAM project (www.dream2020.eu), funded by the European Commission (grant no. 611391)

---

[1] http://www.dream2020.eu/

# Towards an Embodied Simulator of Autistic Child Behaviors: an Improved Method for Selecting Simulated Behaviors

Kim Baraka
Robotics Institute, Carnegie Mellon
University, Pittsburgh, PA, USA
INESC-ID / Instituto Superior Técnico,
Universidade de Lisboa
Lisbon, Portugal
kbaraka@andrew.cmu.edu

Francisco S. Melo
INESC-ID / Instituto Superior Técnico,
Universidade de Lisboa
Lisbon, Portugal
fmelo@inesc-id.pt

Manuela Veloso
Machine Learning Department,
Carnegie Mellon University
Pittsburgh, PA, USA
mmv@cs.cmu.edu

## ABSTRACT

We built a behavioral simulator of structured interactions with autistic children of different severities. Our simulator, named ABASim, uses a behavioral model from the state-of-the-art diagnostic tool for autism, namely the Autism Diagnostic Observation Schedule (ADOS-2). On the other hand, we built a prototype of a customizable robot exhibiting typical autistic behaviors according to a restricted version of the same ADOS-2 model. The robot can be customized to display different severities and forms of autism, and autonomously responds to predefined multimodal stimuli, emulating an interaction with a child with autism. We present here an additional step towards interfacing the ABASim simulator with an embodied agent such as a humanoid robot. We specifically contribute an improved algorithm for selecting behaviors from a dataset according to the features characterizing the simulated child based on conflict resolution between behaviors unlikely to co-occur. This contribution brings us closer to our long-term goal of having a fully embodied simulator of behavioral responses of children with autism of different severities under structured interactions, which may have a number of applications including improved therapist training, novel autism therapy tasks involving robots, and education.

## KEYWORDS

Behavioral simulation, Autism, Social agents

## 1 INTRODUCTION AND BACKGROUND

This works builds upon a series of previous works on simulating behaviors of children with Autism Spectrum Disorders (ASD) of varying severities, in the context of structured interactions used for ASD diagnosis. We foresee several real-world applications to the idea of embodied simulation of ASD behaviors in agents such as humanoid robots. Current therapist training for ASD diagnosis, ignoring the important interactive and embodied component of diagnostic procedures, may greatly benefit from additionally utilizing interactive robots capable of exhibiting typical ASD behaviors in response to standard stimuli. Also, such robots could be used for educational purposes such as classrooms or museums to educate people about potential behavioral differences in individuals with ASD. Last but not least, the ability to simulate expected behaviors of children with ASD has the potential to inform the autonomy of robots used for autism therapy, potentially enabling automated adaptation and personalization mechanisms.

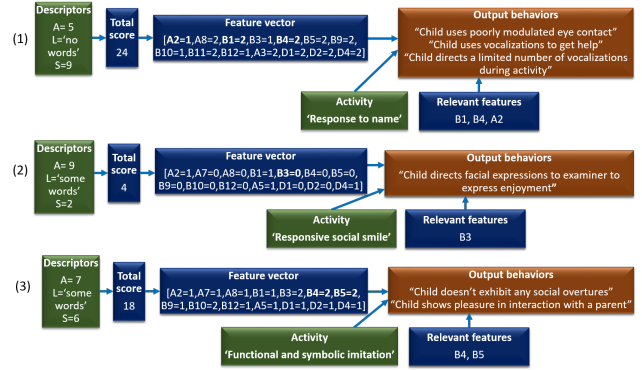

Figure 1: Three examples of ABASim simulating behaviors for specified child descriptors and activity. We show the pipeline from inputs to output, which is based on models from the ADOS-2. For more information, please consult [2].

### 1.1 Simulator overview

We built a simulator, named ABASim [2], based on a behavioral model from the state-of-the-art diagnostic tool, the Autism Diagnostic Observation Schedule (ADOS-2) [3]. It algorithmically reverses the diagnosis pipeline to stochastically output behaviors starting from high-level descriptors of the child, namely the age (A), language ability (L), and ASD severity (S), as illustrated in Fig. 1.

### 1.2 Customizable 'autistic robots': an embodied behavior visualization

In addition, we designed and validated a set of robot behaviors that correspond to behaviors typically observed in children with varying ASD severities. We integrated those behaviors as part of an autonomous agent running on a NAO humanoid robot. Features can be controlled directly by the user and influence the responses of the robot to the different standard stimuli used, inspired by the ADOS-2 activities. Fig. 2 summarizes the architecture of our system.

## 2 METHOD FOR BEHAVIOR SELECTION

In our previous work [2], we output behaviors by first constructing a subset of relevant behaviors, given the feature values to be simulated and the relevant features for a specific activity. From this subset, the algorithm selected behaviors randomly (one for each relevant feature). This method had some issues since it often outputted behaviors that were incompatible, in other words, whose descriptions somehow contradicted themselves. Examples of incompatible behaviors are: "Child exhibits an odd cry and **no other vocalizations**." and "Child **vocalizes** to be friendly", or "Child uses **poorly modulated eye contact** to initiate social interaction." and "Child uses **eye contact** to get help.".

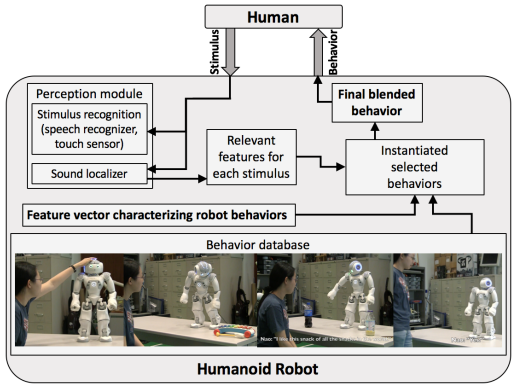

**Figure 2: Architecture of our customizable 'autistic' robots. For more information, please consult [1].**

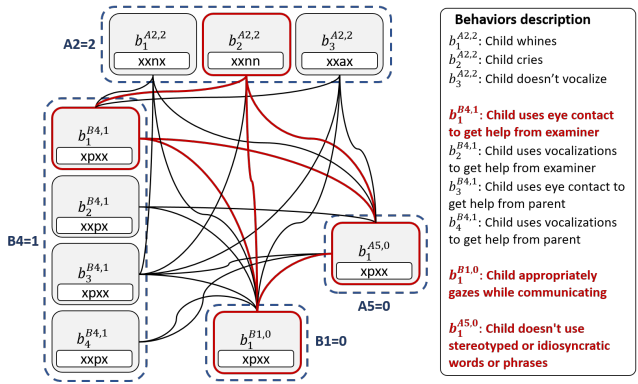

**Figure 3: Sample behavior compatibility graph for instantiated feature values. Behavior $b_i^{f,v}$ corresponds to feature $f$ with value $v$, and indexed $i$. In red: choice of pairwise compatible behaviors.**

## 2.1 Graph representation of behaviors

We introduce a graph representation of behaviors, where nodes represent behaviors and the presence of an edge between two nodes signifies that the two corresponding behaviors are compatible. Due to the large number of behaviors, we cannot possibly define by hand the compatibility of every two pair of behaviors, so we have to automate it. Figure 1 shows a sample graph for four features, namely A2 ('Frequency of vocalizations'), B4 ('Integration of gaze during social overtures'), B1 ('Unusual eye contact'), and A5 ('Stereotyped use of words'), considered relevant for activity 'Response to name'.

## 2.2 Building the behavior compatibility graphs

For each activity and combination of feature values, one can build the compatibility graph containing relevant behaviors as nodes, with edges between compatible nodes that don't belong to the same feature. As an approximate solution for specifying behavior compatibilities, we introduce *behavioral channels*, corresponding to dimensions of social behavior in relation to an embodied agent. The behavioral channels we consider are: Body motion, Gaze, Speech, Emotion/Facial expression, but could differ according to the agent or simulation purpose. On each of those, we define 4 possible values:

- *x*: no mention of specific behavioral content on channel
- *a*: specified absence of behavioral content on channel
- *p*: presence of 'positive' behavioral content on channel
- *n*: presence of 'negative' behavioral content on channel

To each behavior, we associate a behavioral channel vector consisting of four values, one for each channel, listed in the order above under each node in Fig. 3. Compatibilities are determined according to valid combinations of the above values (*n*, *p*, and *a* should not co-occur on the same channel for two compatible behaviors).

## 2.3 Work in progress: Evaluation

We are currently working on evaluating our behavior selection algorithm against a baseline, as well as behaviors observed in real ADOS-2 sessions. The study will involve trained therapists looking at simulated scenarios and answering questions subjectively based on their experience with real scenarios.

## 3 CONCLUSION

This work has presented a step towards a realistic embodied simulator of behaviors of children with ASD in the context of structured interactions, inspired by the ADOS-2 diagnostic tool. In particular, we described an improved method for selecting behaviors from a dataset of possible behaviors, which takes into account compatibility between pairs of behaviors.

In the future, in addition to evaluating our method as well as the feasibility of its numerous applications, we are designing robot-led activities for children with ASD, inspired by the ADOS-2, where we could use the ABASim simulator as part of an agent controlling the robot's actions to adapt to individual differences

## ACKNOWLEDGMENTS

This research was partially supported by the INSIDE project, funded by the CMUP-ERI/HCI/0051/2013 grant, and national funds through Fundação para a Ciência e a Tecnologia (FCT) with reference UID/CEC/50021/2013. The views and conclusions contained in this document are those of the authors only. The authors would like to thank the contributors of the ABASim code, as well as the team at the Child Development Center of the Hospital Garcia de Orta in Almada, Portugal, for their precious collaboration.

# Long-term Robot-Mediated Interventions for Children with Autism: A Case Study

David Becerra
University of Southern California
dbecerra@usc.edu

Maja Matarić
University of Southern California
mataric@usc.edu

## ABSTRACT
Autism spectrum disorders (ASD) are lifelong conditions that affect millions of people worldwide. Recently, researchers have explored the promising utility of socially assistive robotics (SAR) as a therapy tool to help people, particularly children, with autism. However, research has rarely considered the dynamics of robot-mediated autism interventions in long-term studies. In this paper, we present a long-term in-home intervention of a socially assistive robot for children with autism. We provide an overview of the robotic system as well as the study design of the intervention. Finally, we present a case study of a child with ASD and discuss our preliminary results.

## KEYWORDS
Human-robot interaction, socially assistive robotics, autism spectrum disorders, long-term, in-home

## 1 INTRODUCTION

In the United States alone, more than 3 million people live with autism spectrum disorders (ASD), a range of developmental conditions often characterized by atypical social behaviors [1]. In an effort to support and expand the often costly resources available for individuals with ASD, a growing area of research is exploring the potential for socially assistive robotics (SAR) as a therapeutic tool for those diagnosed with autism [5].

Socially assistive robotics is a nascent field of computer science aimed at understanding how robots can help people via social rather than physical interactions [3]. In autism research, SAR has already shown promise in eliciting novel behaviors such as turn-taking and joint-attention [2, 4]. However, in order to fully understand the impact and usefulness of SAR for people with autism, longer-term interventions must be conducted.

In this paper, we present a long-term robot-mediated intervention exploring how socially assistive robots can promote the cognitive and academic development of children with autism. First, we discuss the system and study design of the in-home intervention. Next, we present a case study of the intervention with a single child with ASD. Lastly, we present preliminary results from the case study and discuss future research plans.

## 2 APPROACH

To investigate how SAR can be leveraged to assist children with ASD during a long-term intervention, we developed an autonomous robotic system that can be safely deployed into homes over extended periods of time. Any in-home intervention should be mindful of the privacy and comfort of its patients. To that end, our system is designed to be minimally intrusive and maximally accessibile.

### 2.1 System Design
At the center of the system is the SPRITE robot Kiwi, a Stewart-Platform designed for research in socially assistive robotics [6]. Children interact with Kiwi by playing 10 different educational games on a touch-screen monitor. The games, designed around a space theme, explore numeracy concepts such as counting, ordering/sequencing, and pattern matching. Each game has five difficulty levels, with higher difficulties requiring more advanced skills such as simple mathematics. All of the games were developed with educational specialists to ensure the content was appropriate for our target age range, children 4-6 years old.

While playing the games, Kiwi behaves autonomously. It delivers instructions for the child to complete each exercise, while also providing feedback and support whenever the child needs help or makes a mistake. All of Kiwi's dialogue and gestures are triggered automatically by events from the game.

As stated previously, the system is designed to be accessible, particularly for families with minimal technical experience. To begin interacting with the system, the family must physically turn on the computer. Once on, all the software including the games and all data recording is automatically activated. Similarly, the family must flip a physical switch to turn on the robot. No further steps are required from the family to start the system.

### 2.2 Study Design
The socially assistive robot system is part of an ongoing within-subjects user study involving typically developing children and children with autism. In the study, each participating family has the robot in their home for 30 days. During the duration of the study, participants are asked to interact with the system in 20 separate sessions, at most one session on any given day. A session lasts approximately 10-25 minutes; however, we allow participants to play as long or as short as desired.

During a single session, the child plays the ten games with Kiwi while being audio- and video-recorded. Video recordings capture a front view of the child as well as anyone else directly behind the child. The audio and video recording automatically stop whenever the family turns off the system.

After the child finishes the games, the parent completes a short interaction questionnaire that contains the following mix of 5-point Likert scales and open-ended questions: (1) How did your child feel about interacting with the robot today? (2) To what degree did you have to motivate your child to interact with the robot and games today? (3) What did you use to motivative your child today? (4) How independently did your child interact with the robot and games today? (5) How do you think your child performed on the activities today? (6) Which of the following best describe your child's mood or feelings today? (7) Did anything unusual take place in your child's day today? (8) In a few words please describe how your child is developing in terms of learning so far? (9) In a few words, please describe how your child is interacting and engaging with the robot so far.

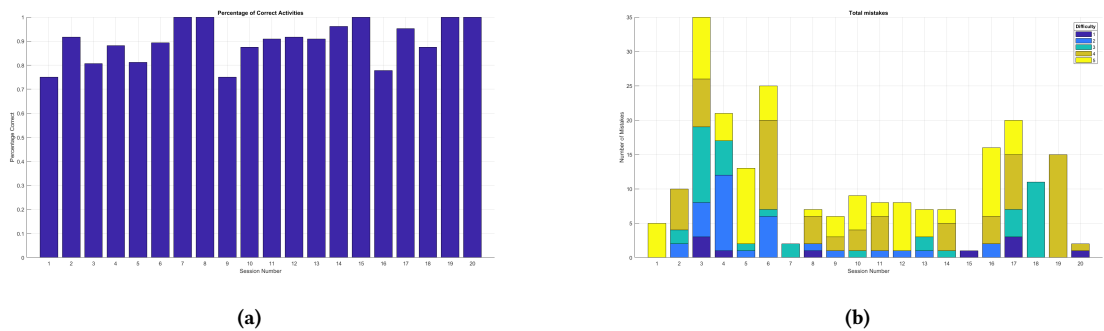

**Figure 1:** (a) Child participant's performance across all 20 sessions. (b) Total mistakes per difficulty level the child made across all 20 sessions.

Additionally, every week the parent participates in an informal semi-structured interview as an opportunity to provide a brief status update as well as to communicate any technical issues that may have arisen. At the end of the 30 days, the system is collected from the participant's home and an informal post-study interview is conducted.

## 3 PRELIMINARY RESULTS

Rigorously analyzing the results of this ongoing study is premature and beyond the scope of this paper. However, we present preliminary results from a single deployment as a case study of the capabilities of SAR for long-term interventions for children with ASD. At the time of this publication, a total of two deployments have been conducted. We choose to focus our attention on the second deployment as several technical improvements were made following feedback from the initial deployment.

The particular case study discussed in this paper involved a 7-year old male child with ASD. The child had the robot in his home for a total of 37 days, during which time he interacted with the robot 20 times. Each session lasted 10.8 minutes on average ($SD$ = 7.6 minutes). The initial analysis of the child's performance on the games has not yet shown any significant results. Figure 1 shows that the child consistently answered 70% or more games correctly ($mean$ = 89.9%, $SD$ = 8.4%); there appears to be no significant change of performance overtime. In addition, we note that the child makes significantly more mistakes on the higher difficulty problems, with 64% of mistakes happening at difficulties 4 and 5.

We observed encouraging behavioral changes in the child's interaction with the robot system over time. On the day researchers brought the setup to the home, the child was hesitant to interact with the robot or play the games. However, as the study progressed, we observed that the child was not only excited to play with Kiwi, but he actively told his peers about the games and invited relatives to play alongside him. Similar results were noted in the semi-structured weekly interviews. The parent expressed that the child was initially scared of Kiwi. However, as the parent gradually exposed the child to the robot, the child grew fond of Kiwi and wanted to share it with his friends, therapist, and social worker. These observational results suggest the potential for long-term interventions to increase human-robot attachment.

Furthermore, initial analysis of the parent survey indicates an increase in the child's autonomy while playing the games. During the first few sessions, the parent reported that the child required a moderate amount of assistance while playing the games. However, during the final sessions of the study, the parent began reporting that the child was able to complete the games independently, requiring little to no assistance. These initial reports indicate the possibility for SAR interventions to help a learner with autism gain academic autonomy and independence.

## 4 FUTURE WORK

We plan to continue to analyze the data collected from the described case study. We are annotating and investigating the video recordings to explore how features such as eye gaze and speech can illuminate how attentive and engaged the child was throughout each interaction. As this is part of a larger study, we are continuing to deploy the system into additional homes, and making systematic improvements along the way. Once the study is completed, and data analyzed, we will publish the results as a longitudinal study of SAR's promising applications to autism therapy.

## 5 ACKNOWLEDGEMENTS

This research is supported by the NSF Expeditions in Computing for "Socially Assistive Robotics" (IIS-1139148).

# Robotic Therapies: Notes on Governance

Eduard Fosch-Villaronga[a] and Jordi Albo-Canals[b]

[a] *Microsoft Cloud Computing Research Center & Queen Mary University of London, United Kingdom*
[b] *Research Affiliate, Center for Engineering Education Outreach (CEEO) at Tufts University, United States*

*Abstract*— The insertion of robotic and artificial intelligent systems in therapeutic settings is accelerating. This paper investigates the legal, ethical and societal challenges of the growing inclusion of cyber-physical systems that are used for therapeutic purposes, and explores solutions. Typical examples of such systems are Kaspar, Hookie, Pleo, Tito, Robota, Leka or Keepon. As these technological developments interact socially with children, elderly or disabled, these may raise concerns other than mere physical safety compromise, including data protection, inappropriate use of emotions, invasion of privacy, autonomy suppression, human-human interaction decrease or cognitive safety. Due to the novelty of these technologies and the uncertainties of the impacts of their applications, it is neither clear what is the appropriate applicable legislative framework. Our contribution seeks to provide a thoughtful and thorough analysis of some of the most pressing legal and regulatory issues in relation to cyber-physical systems used for therapeutic purposes, in the hope that this will inform the policy debate and set the scene for further research.

*Keywords*— Therapeutic Robots, Ethical Legal and Societal Issues (ELSI), Personalized Care, Emotion Regulation, Privacy, Evidence-based Policies, Guidelines.

## INTRODUCTION

The overall objective of this project is to explore legal and regulatory challenges in regulating therapeutic robot and artificial intelligence technologies and explore potential solutions.

There are currently legal and regulatory initiatives governing robots, users and roboticists, including robot private standards - ISO 13482:2014 Personal Care Robots, BS 8611:2016 Guide to the ethical design and application of robots and robotic systems, and IEEE Ethically Aligned Design 2017 from the IEEE Global Initiative and Standard Association - or public policymaking, including European regulatory initiatives (Resolution 2015/2103 (INL) 2017 and its response from the European Commission) and international public policies on drones, self-driving cars and delivery robot regulations (mainly in the United States). Even so, these initiatives or in-force laws may not give an adequate response to those roboticists working in different domains of application, e.g. on robots that interact socially with humans for therapeutic or educational purposes.

This short paper provides a glimpse of concrete legal and regulatory issues in relation to cyber-physical systems in therapeutic settings, which is something novel in comparison to the general guidelines for robot governance that have been released so far (Robolaw Project 2014, RockEU 2016, AI Now Report 2017). In the full paper, we will extend such analysis and will devise specific-domain and robot-type sensitive policy recommendations for therapy in the hope that this will inform the current policy debate.

## RELATED WORK

A review of the literature reveals an emphasis on the benefits of robots for cognitive therapies. In general, robots can meet the specific needs of a disease either at a physical (exoskeletons) or at a cognitive level in a personalized way, they can useful as a diagnostic aid, they can serve as behavior eliciting agent, or social mediator (Cabibihan 2013). Furthermore, robot's behavior is predictive and repetitive, and less complex or intimidating than humans, which can be beneficial in different contexts, e.g. for autistic children.

Despite these benefits, robot technology may cause unexpected moral implications (Salem et al., 2015). In the case of social robots interacting with children under the autism spectrum disorder, for example, relevant literature apprises some of these implications, including acceptability, trust, sociability, or attachment issues (Coeckelbergh et al. 2015). To this regard, the European Parliament (EP) warns that the incremental use of care robots could lead in the future to the dehumanization of health practices.

Although a bit alarmistic because the majority of robots in therapy are used as a tool, as an extension (not replacement) of the therapist, little is known about the possible risks of current and future uses and developments of therapeutic robots. This may be due to the novelty of practices, the lack of interdisciplinary research, the current physical-safety focus of available standards, or a benefit-centered research approach. The fact that there are still no general and accepted quantitative methods to evaluate such therapies (Scassellati, Admoni, Matarić 2012), and that robot therapies are not yet mainstream in healthcare sector do not make the discernment of these concerns clearer either.

## PRELIMINARY FINDINGS

The following subsection briefly compiles, in italics, the legal and ethical challenges that we have identified so far regarding therapeutic robots and artificial intelligence technologies. Each concern is accompanied by a short note that contextualizes and explains them. This investigation has been mainly based on literature research. The appraisal of the identified issues is built on legal and regulatory analyses from previous related work, in particular concerning recent European projects (Robolaw Project 2014, RockEU 2016) and other relevant reports (AI Now Report 2017, Intel Report 2017). The findings of project contribute to these analyses by exploring recent private settings that govern the safety design of robots (ISO 13482:2014, BS 8611:2016, IEEE EAD 2017), public policies governing other types of robots (drone, self-driving car and delivery robot), and the latest European regulatory initiatives (Resolution 2015/2103 (INL) 2017 and its response from the European Commission). The second subsection concisely outlines our proposed solutions to the arisen challenges.

### a. Identified Legal and Ethical Challenges

- *Disagreements in the technical literature impedes clear discernment of the importance of associated ethical, legal and societal issues (ELSI).* While some researchers argue that by allowing the robot to show attention, care and concern for the user (Tucker 2006) as well as to being able to engage in genuine meaningful interactions, socially assistive robots can be useful as therapeutic tools (Shukla et al. 2015); other studies suggest that, actually, the emotional sharing from the robot to the user does not necessarily imply feeling closer to the robot (Petisca et al. 2015). This impedes the production of common/agreed guidelines framing the use and development of these robots.

- *The lack of standardized procedures and guidelines impedes the establishment of a clear safeguard baseline for therapeutic robots.* In order to deliver effective care, healthcare settings use standardized clinical reasoning terms, evidence-based assessment criteria for selecting appropriate diagnoses, activities for interventions and indicators for different outcomes. While animal assisted therapy is recognized as a nursing intervention under the North American Nursing Diagnosis Association (NANDA), (pet) robot (assisted) therapies do not enjoy the same category (Alvaro-Rodero and Garcia-Fernandez 2016). This affects the understanding of what requirements make 'safe' a robotic therapy. This may also hinder public access to these therapies and widen the social divide.

- (For instance) *It is uncertain whether the use of emotions in HRI affects the user in robot therapies.* At this stage it is not clear under which circumstances or in which contexts robots can collect and process emotions of the users, and interact emotionally with them. We do not know whether such practice could challenge the rights of the user in the near or long term (Richardson 2018), and whether the law should appropriately address it - probably by establishing a purpose limitation. In this respect, the inappropriate use of emotions could challenge several rights, from privacy to autonomy, especially because 'emotions and decision-making go hand in hand' (Lerner et al. 2015).

- (in similar line) *There are no safeguards for cognitive human-robot interaction*. The cognitive aspects concerning HRI - perceived safety (Salem et al. 2015) or psychological harm - are often disregarded, while we do interact with robots on the cognitive layer (Fong, Nourbakhsh, and Dautenhahn 2003). Because the law protects both the physical and the psychical integrity of the user, more research is needed to understand (and maybe develop) cognitive safeguards for safe HRIs.

- *Disobedient and imperfect robots that enhance long-term engagement may risk legal compliance*. In order to establish relationships and permanent attachment between artificial social agents and humans, social entities require personality. To make it more real or alive, personality is created through unique imperfect behaviors. The robots may end up being unique, disobedient and imperfect (Konok et al. 2018). This may compromise the predictive behavior of the robot, something crucial in some therapies – e.g. for autism (Dautenhahn 2000). This affects the 'safety' of the therapy, the compliance certification processes (because every robot is unique) and the trust and reliability of the user.

- (moreover) *Depending on the type of HRI and the robot embodiment, several dimensions of privacy can be affected*. These include bodily, spatial, communicational, proprietary, intellectual, decisional, associational, behavioral and informational (Koops et al. 2016).

- *Upcoming data protection principles may have to be respected*. Although it is not clear whether 'emotions' are mere biometrical data (and thus sensitive data with a higher degree of protection) or whether, in the future, they could become a standalone category within the concept of 'personal data;' the European General Data Protection Regulation (GDPR) will apply. This corpus includes undefined concepts such as privacy-by-design, right to be forgotten or the right not to be subject to automated decision-making. Although the right to explanation is not a right under such corpus per se, recital 71 GDPR and relevant literature suggest that the more and more robots will have to be able to explain their actions (Johnson 2015). Who should therapy robots report to?

- *Various emerging legal concepts and regulatory models that aim at mitigating risks relating to cyber-physical systems (kill switch, reversibility or responsibility of the teacher) risk at being ineffective for cognitive therapeutic robots*. Regulatory initiatives that fail to address the inter-dependence of the tangible and virtual elements of cyber-physical systems risk being ineffective (Fosch-Villaronga and Millard 2018). For instance, a physical action of the robot may be irreversible; there exists no protective stop (or red button) for cognitive processes or data collection, and the 'teacher' of the robot can be another robot that uploaded the behavior to the cloud.

- *There are no codes of conduct for roboticists*. If HRIs integrate social behaviors, those who design, use and control this kind of robots should also be required moral agency and emotion (Coeckelbergh 2010). Similarly, the EP pointed out that 'robotics engineers should remain accountable for the social, environmental and human health impacts that robotics may impose on present and future generations.' The question is: who will write those codes? How are they going to be enforced?

- *The cyber-physical duality of robots may obscure the origin of a particular problem, the scope of its consequences, and its subsequent impacts in the two 'worlds'*. This duality complicates legal relationships and liabilities among various actors, such as users, manufacturers, and cloud service providers, which is exacerbated by the growing existence of Robot-as-a-Service practices (Fosch-Villaronga and Millard 2018).

  b. *Proposed solutions to the arisen challenges*

- *The collection of empirical data from several research projects could help draft international guidelines to promote safer, richer and more effective therapies*. This could push authorities – e.g. NANDA – towards the inclusion of robotic therapies within standardized interventions (Barco and Fosch-Villaronga 2017).

- *The development of new principles may help mitigate compounding risks and shape future robotic therapies framework*. These principles may include: non-isolation (human-human interaction), individualized-care (personalized care), no-assumption (value sensitive and user-centric design), policy-learning (evidence-based), and accessibility (low-cost and minimal design).

- *The inclusion of codes of conduct and standards in private contracts may help the enforcement of these soft law instruments*.

- *A complete data protection and privacy impact assessment could adequately assess and mitigate related risks*.

## CONCLUSIONS & FUTURE WORK

The insertion of robots and artificial intelligent technologies in therapeutic settings raise challenges at the legal and ethical level. Although several regulatory initiatives have been released, we argue that they may not suffice to appropriately mitigate the specific risk that these robots present in this specific domain of application. Future work will include a thorough explanation of the legal and ethical challenges and we will explore the solutions in detail.

## ACKNOWLEDGEMENTS

This paper has been produced by a member of the Cloud Legal Project, Centre for Commercial Law Studies, Queen Mary University of London. The author is grateful to Microsoft for the generous financial support that has made this project possible. Responsibility for the views expressed, however, remains with the author.

# Social Robots as Alternative Non-Pharmacological Techniques for procedures in children

Gloria Beraldo
Department of Information
Engineering, University of Padua
Italy

Emanuele Menegatti
Department of Information
Engineering, University of Padua
Italy
emg@dei.unipd.it

Valentina De Tommasi
Terapia Antalgica e Cure Palliative
Pediatriche, UOSD of Padua
Italy
casadelbambino@aopd.veneto.it

Roberto Mancin
Dipartimento di Salute della Donna
e del Bambino, University of Padua
Italy

Franca Benini
Terapia Antalgica e Cure Palliative
Pediatriche, UOSD of Padua
Italy

## ABSTRACT
In pediatric hospitals children undergo several medical procedures, many of which are often painful, unexpected and leads situational stress and anxiety. In this paper, we describe a pilot study to inquire the possibility of introducing social robots as alternative non-pharmacological techniques for procedures in children. The robot could entertain the patients, interact physically and verbally with them, in order to decrease their fear, manage their pain, relax and give them the best support. In this study, we exploit two different social robots, Pepper Robot and Sanbot Elf, for the purpose of understanding how they are perceived by patients and what features of each robot impact more on the children.

## CCS CONCEPTS
• **Human-centered computing** → **Empirical studies in HCI**;

## KEYWORDS
Social Robots, Pediatric, Human Robot Interaction

**ACM Reference Format:**
Gloria Beraldo, Emanuele Menegatti, Valentina De Tommasi, Roberto Mancin, and Franca Benini. 2018. Social Robots as Alternative Non-Pharmacological Techniques for procedures in children. In *Proceedings of Workshop on Social Robots in therapy: focusing on autonomy and Ethical Challenges (SREC18)*. ACM, New York, NY, USA, Article X, 2 pages.

## 1 INTRODUCTION
Children often have to experience painful and invasive procedures that might impact negatively on their quality of life: they may have long-term consequences on behaviour, memory, pain perception and their development [5]. In addition, it is complex to understand how children patients feel pain, because it entails analysing physiological, psychological, behavioral and developmental factors [8]. As well as the pain, fear and unfamiliar surrounding can lead to a high level of anxiety, which may complicate the performance of the procedure. Indeed, when anxiety is alleviated, children are more cooperative, likely to engage with clinicians and follow instructions [3]. For these reasons, sedoanalgesia has been often used to sedate patients for a variety of pediatric procedures [6]. In the process of sedoanalgesia exploiting non-pharmacological techniques is crucial and complementary to take medicines [4]. The most common traditional non-pharmacological approaches consist of psychological preparation and information, passive and active distraction, relaxation techniques, music therapy, guided imagery and hypnosis [9].

During the last and the current years, some researchers are investigating how social robots could complement the service and the support provided by human specialists. For example, Beran et al. examined the effectiveness of NAO robot in pediatric hospital to implement cognitive-behavioral strategies while children received a flu vaccination [1]. In [2] Jeong et al. showed that children are more emotionally engaged in interacting with a robot than a virtual character on a screen during patients intervention session.

The key research question of this paper is to inquiry the possibility of introducing social robots as alternative non-pharmacological techniques for procedures in children, without becoming a replacement for every members of the staff. This research extends the work conducted at *Terapia Antalgica e Cure Palliative Pediatriche* of UOSD in Padua, in which NAO robot was used to manage the anxiety and the fear of the children before sedation and analgesia [7].

## 2 MATERIALS AND METHODS
This study provides the use of two different types of robots shown in Figure 1: (i) Pepper by Softbank Robotics, a humanoid robot created to communicate with people in the most natural way, through its body movements and voice, and (ii) Sanbot Elf by Qihan, able to display facial expressions, speak, animate arms motions combined possibly with music, dance and different colors flashing LEDs. The choice

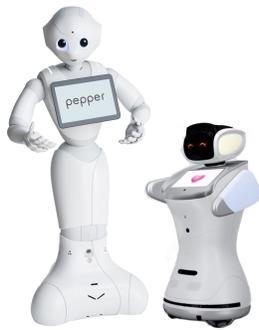

Figure 1:
(left) Pepper Robot
(right) Sanbot Elf

of robots was due to the availability at UOSD in Padua. The hypotheses of this study are as follows:

*H1: The effectiveness of using social robots as alternative non-pharmacological techniques for procedure in children will be determined by how the robots will engage the patients.*

*H2: The kind of robot will be preferred by children depends on their age, gender and personality and also appearance of the robot.*

### 2.1 Participants

Since the difficulty to recruit children whose parents accept to introduce robots during procedures and sometimes the little collaboration in filling in the questionnaires, we are scheduling 26 participants in order to obtain a statistical power equal to 0.80 with Cohen's d of 0.8. Our inclusion criteria provide all patients aged 4-18 stay at Clinica Pediatrica of Padua, whose parents will sign the consent form. The participants are divided into two groups balanced according to age, gender, nationality and other social conditions: one will interact with Pepper Robot, the other with Sanbot Elf.

### 2.2 Procedure and Questionnaire

First of all, the child patient is informed by the assigned health care researcher that a specific robot (Sanbot or Pepper according to the list with patient-robot assignments) is waiting for him/her in the room of the procedures. Then the corresponding robot welcomes the child and it introduces itself. It interacts with the patient by conversing about their likes/dislikes, dancing, moving the arms, playing guessing games according to the specific child. The robot was operated in Wizard of Oz (WoZ) mode by an operator, using text-to-speech synthesis (TTS) and gestures. The interaction lasts for few minutes until the sedation kicks in. At the end researcher gives a sheet to the patient's parent, with the picture of the robot, to thank the child for playing with it and to inform him/her that when he/she wakes up, he/she will meet again it.

The patient's parents will be ask to compile a questionnaire, specifically designed for evaluating the perception of anxiety, fear and pain before and after the procedure. In particular we are interested in the strong points and (eventual) shortcoming of the two robots used in this study. All data will be anonymous and collected with respect for individual privacy. To analyse our data we will conduct a MannWhitney U test.

## 3 CONCLUSIONS & FUTURE WORKS

In this paper, we propose a protocol for a pilot and qualitative study to investigate the possibility of introducing social robots as alternative non-pharmacological techniques for procedures in children. Through the questionnaire, we will try to figure out the kind of emotions felt by the patients before the procedure and if they are changed or not after the interaction with the robot. Furthermore, we will attempt to understand how the children patients will be helped by the robot to reduce their anxiety and fear and what features of the robots impact more to engage them.

In future work we plan to deepen this research taking into account also the other social actors involved and sustaining it by other devices in order to achieve quantitative measures. For example, it would be very useful to exploit cameras to analyse the patient's facial expression, the number and the kind of their movements and to collect fruitful data possibly useful to make the robot more autonomous. Furthermore it could be possible to evaluate more precisely the emotional condition through electrodermal activity (EDA) sensors.

## ACKNOWLEDGMENTS

This research was partially supported by Fondazione Salus Pueri with a grant from "Progetto Sociale 2016" by Fondazione CARIPARO. We thank Andrea Sattanino and Isabella Mariani, neuro-psychometrician and psychology undergraduates, for their help to run the future experiments. We would also like to show our gratitude to the *Terapia Antalgica e Cure Palliative Pediatriche* at UOSD of Padua.

# User Expectations of Privacy in Robot Assisted Therapy

Zachary Henkel, Kenna Baugus, Cindy L. Bethel
Mississippi State University
Mississippi State, Mississippi
zmh68,kbb269,clb821@msstate.edu

David C. May
Mississippi State University
Mississippi State, Mississippi
dmay@soc.msstate.edu

## ABSTRACT

This article describes ethical issues related to the design and use of social robots in psychological interventions and provides insights from two user studies. Expectations regarding privacy with a therapeutic companion robot gathered from a 16 participant design study are presented. Additionally, perceptions and beliefs related to privacy and social judgment from 87 child participants following an interview related to sensitive content are reviewed. These findings demonstrate the need for further investigation into the expectations and beliefs surrounding the use of therapeutic robots.

## CCS CONCEPTS

• **Security and privacy** → **Privacy protections**; **Social aspects of security and privacy**; **Usability in security and privacy**;

## KEYWORDS

robot assisted therapy, privacy, ethical robot design

## 1 INTRODUCTION

As robotic systems become more reliable and autonomous, their use in therapeutic applications presents many unique challenges. Our ongoing research includes the design of a therapeutic robotic dog, Therabot$^{TM}$, and the development of social robots for conducting forensic interviews with children. The development and use of these systems presents ethical questions, which must be incorporated into the research process [2]. In this article we focus on the question: *How do user expectations and beliefs about autonomous social robots for psychological interventions affect ethical usage?*

## 2 BACKGROUND AND APPROACH

As the research community has grappled with ethical questions relating to robot-assisted therapy, many have arrived at mechanisms for limiting or augmenting full autonomy [1, 3]. In addition to developing standards and recommendations, which advise clinicians and patients on the usage of social robots, it is also helpful to consider the beliefs and expectations that users possess about the robots that will assist them.

As our research developing a therapeutic robotic dog continues, we have collected data from potential users related to their expectations with regards to privacy. Concurrently, our research efforts in the area of using social robots as forensic interviewers for children have led us to further investigate the beliefs children hold about social robots they interact with. The two studies presented in this article investigate these topics through the use of semi-structured verbal interviews with a researcher. The resulting qualitative data has been coded into discrete categories and distilled into areas that merit further investigation.

## 3 EXPERIMENTS

The studies involved in the development of Therabot$^{TM}$ and our forensic interviewing platforms have captured insights into the beliefs and expectations of potential users for each system. In this section we present responses from adult participants related to privacy while interacting with Therabot$^{TM}$ and responses from children related to privacy and trust following a forensic interview about their experiences with bullying with a human or robot interviewer.

### 3.1 Therapeutic Robot Design Study

As part of the development of the therapeutic robotic dog, Therabot$^{TM}$, participants took part in a collaborative design process consisting of surveys and a semi-structured interview with a researcher concerning the potential uses of Therabot$^{TM}$. During the session participants were free to demonstrate their ideas using words, drawings, and physical prototypes.

Participants were specifically asked how willing they would be to let the robot record their conversations with their therapist during therapy sessions, as well what should be taken into consideration with regards to protecting a user's privacy.

Sixteen participants (9 female, 7 male) recruited from the university's psychology program completed the study. A majority of participants (14/16, 88%) reported that audio recording of therapy sessions or user activity data would be acceptable under certain conditions. Participant provided restrictions included: obtaining consent from users (6/16, 38%), recording only data useful for therapy (3/16, 19%), and limiting recording to therapy sessions (3/16, 19%). Some discussed features such as indicator lights informing users of data recording, while others focused on limiting who would have access to the recorded data.

Questionnaires included a series of questions asking participants to rate *How harmful or beneficial would an interactive (robotic) comfort object be for each age group?* for the age groups: *Children (0-11), Adolescents (12-17), Adults (18-59), and Older Adults (60+).*

Table 1 shows the mean and standard deviations for each age group. A Maulchy's Sphericity Test indicated data did not violate assumptions of sphericity ($\chi^2(5) = 8.31, p = 0.141$). A significant difference was found between the helpfulness rating for each age group ($F(3, 42) = 6.16, p = .001$). Post-hoc pairwise comparisons indicated significant differences between Children and Adults ($p = .009$).

| Age Group | Mean | SD |
|---|---|---|
| Children (0-11) | 6.27 | 1.03 |
| Adolescents (12-17) | 5.40 | 0.63 |
| Adults (18-59) | 4.73 | 0.88 |
| Older Adults (60+) | 5.47 | 1.25 |

**Table 1: Harmful/Helpful ratings for each age group, on a 1 (very harmful) to 7 (very beneficial) scale.**

Though from a small sample of 16 participants, these responses indicate that the therapeutic platform was thought to be most useful for children, a population often afforded unique protections. This indicates that the beliefs and perceptions of child users are likely to be essential to answering questions concerning ethical usage of the platform.

### 3.2 Interviewing Children About Sensitive Topics

As part of an effort exploring the use of social robots in forensic interview scenarios with children, we conducted a study focused on bullying experiences at school with children (8-12 years old). Participants engaged in a thirty minute forensic interview with either a human male, human female, male robot, or female robot acting as the interviewer. Humanoid Robokind R25 robots were used for the robot conditions. Following the interview, a researcher verbally administered an open-ended survey to better understand the participants' perception of the interviewer. Sessions were recorded with parental consent and participant assent.

We examined transcripts of the participants' verbal responses to the perceptions of the interviewer survey and identified any spontaneous statements related to privacy, social judgment, trust, focused attention, and concern, or knowledge/experience. As shown in Figure 1, significantly more ($\chi^2(1) = 8.49, p = 0.004$) of the participants interviewed by a robot discussed these factors (19/39, 49%) in comparison to those interviewed by a human (38/47, 19%).

Specifically, participants reported beliefs that the robot interviewer would not share the information they discussed with it, that it would not judge them in the same way that other humans would, that it was more focused and concerned about them in comparison to humans, or that they could trust the robot with sensitive information. Some of the responses included:

- "sometimes humans um that when you tell them stuff like that they'll just use it against you and I don't think he'd do that."
- "like personal secrets that I wouldn't tell anyone something like that, like if I had a Hannah in my room I would tell her everything"
- "I felt like comfortable with what he understood and like wouldn't like judge me"
- "I think a lot of other kids have probably talked to her about some of their problems, and she'd probably have some good

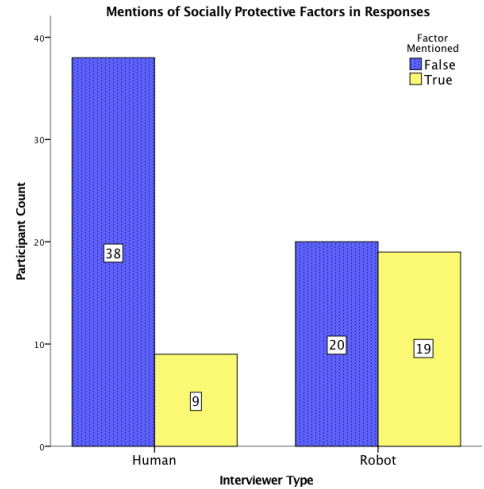

**Figure 1: Socially protective factors discussed by children following interviews about sensitive content.**

ideas about how to stop their problems, and from the knowledge she has from that she'd probably give me some knowledge"

Additionally, in responses discussing socially protective factors, two of those in the human condition presented these factors in a negative context (e.g., "If I tell someone something they might tell other people"), while those in the robot condition only discussed these factors in a positive way.

These findings indicate that many of the children (19/39, 48.72%) felt confident in the robotic platform's ability to maintain privacy and provide them helpful advice. This raises ethical questions related to how an autonomous companion robot should share information and how the user of the robot should be informed about this sharing.

## 4 FUTURE WORK

As we continue to investigate robot assisted therapy and move towards the use of more autonomous systems, ethical practices merit further focused inquiry. Specifically, questions related to how user characteristics shape expectations and beliefs about therapeutic systems should be examined and incorporated into the design process.

## ACKNOWLEDGMENTS

This research was sponsored by NSF Award #1249488 and NSF Award #1408672.

# Toward a taxonomy of social robots for aging-in-place

Lucile Dupuy
Human Factors and Aging Laboratory
University of Illinois at Urbana-Champaign
USA
ldupuy@illinois.edu

Wendy A. Rogers
College of Applied Health Sciences
University of Illinois at Urbana-Champaign
USA
wendyr@illinois.edu

## ABSTRACT
Among the technological solutions to promote aging-in-place, robotics and in particular social robots, have the potential to support this need. However, as an emerging field, there is a challenge to circumscribe the existing concepts and definitions of social robots for this particular topic. Therefore, our work aims at covering the different research fields related to social robots for aging-in-place, and resulting with a framework in the form of a taxonomy. Moreover, our taxonomy highlights the remaining challenges related to the conception and validation of these robots, providing fruitful research avenues for the promotion of older adults' independence at home.

## KEYWORDS
Social robots, aging-in-place, conceptual framework

## 1 INTRODUCTION

Our population is aging. Trends suggest that in 2050, 21.5% of the worldwide population will be aged 60 and over [1]. One of the key challenges related to this aging of the population is the promotion of aging-in-place, wished from both the society and older adults themselves [2]. Among other technological solutions, social robots have the potential to support this need [3]. Nonetheless, as an emerging topic, there is a challenge of defining and circumscribing this object of study. Across research studies, several terms have been used, either as synonyms or as distinct research topics: *social robots, sociable robots, socially assistive robots, assistive robots, therapeutic robots,* among others. An insightful example concerns studies with the harp seal PARO [4], specifically designed for older adults. Among research articles, this robot has been referred by several terms including "artificial pet", "sociable robot", "therapeutic socially assistive pet robot", "social commitment robot", "mental commitment robot", and "companion robot".
Therefore, our work aims to disentangle the different concepts in robotics related to social robots for older adults, and to propose a conceptual framework embracing all these research areas.

## 2 TAXONOMY OF SOCIAL ROBOTS FOR AGING-IN-PLACE

We propose that social robots for aging-in-place lie at the intersection of three broad fields of robotics: social robots, assistive robots, and domestic robots. By building upon existing concepts and definitions related to social robots, we introduce a new taxonomy of social robots for aging-in-place.

Particular details will be given for robots belonging to several categories at the same time. This taxonomy is not specific for robots for older adults, but it is of particular relevance for the support of aging-in-place.

### 2.1 Domestic Robots
A domestic robot is a robot acting in domestic environment for everyday activities, usually by untrained users. Examples of such robots are domestic chore robots, robots for surveillance, logistic robots, telepresence robots, and companion and edutainment robots [5]. These robots have a great potential for supporting aging-in-place as they have to be able to act in every household, and to interact with various users or groups [6].

### 2.2 Social Robots
Social robotics is a very flourishing and broad area of robotics. This research topic focuses on robots able to interact and communicate in an appropriate way with other social agents, namely other robots and humans [7]. Several terms have been used to categorize social robots, depending on the sophistication of its social intelligence, from *socially receptive robots*, relying on human tendency to anthropomorphize (i.e., attribute human characteristics to inanimate objects to rationalize their actions), to *sociable robots*, imitating human social psychology and cognition [8]. Robot's social skills enable to facilitate interaction by using more natural input modalities (e.g., voice, gestures) [9] and provide social presence [8], which is particularly relevant for older adults.

### 2.3 Assistive Robots
Another field of robotics is defined as assistive robotics, focusing on robots giving aid or support to users with special needs such as children, older adults and people with cognitive or physical disability [10]. These robots include rehabilitation robots, wheelchair robots, companion robots, and educational robots, and have the potential to act in different environments (e.g., hospitals, schools, homes). Assistive robots are a promising technological solution to support aging-in-place as their embodiment allows at the same time to assist from basic and physical activities of daily living (e.g., robotic mobility aid, manipulator robots) to more complex and cognitively demanding activities (e.g., medication or activity reminders) [14].

Globally, we can see that variety of the mentioned robots cross-cut these three broad fields, even more when focusing on robots supporting aging-in-place.

## 2.4 Robots for aging-in-place

The following definitions are mainly inspired by existing ones, but are here detailed in the light of the three categories of robots previously mentioned: domestic, social and assistive robots.

A **homecare robot** is a robot acting in domestic settings dedicated to users with special needs [15]. Example of homecare robots are robots fetching and carrying objects to human users. They may behave without interacting socially with the user (e.g., PR2) or have some social skills to facilitate interaction (e.g., Care-O-bot).

An **assistive social robot** is a robot having both social skills and an assistive purpose [13]. Examples of assistive social robots are coach robots (e.g., Bandit) or educative robots. They may act at home or in other environments (e.g., schools, hospitals).

Finally, **companion robots** are robots behaving in domestic settings and providing companionship thanks to some social skills [14]. They may or may not have an assistive purpose. Examples of companion robots are pet robots (e.g., AIBO, PARO) or informative robots (e.g., Jibo).

It has to be noted that depending on the purpose, the same robot can belong to a different category. For example, if the Paro is used in therapy for older adults with dementia, it would belong to the assistive social category, whereas if it is deployed in the home of healthy older adults to provide social presence, it would be considered as a companion robot.

## 3   CONCLUSIONS

Based on existing literature, we introduced a new taxonomy of social robots for aging-in-place. We highlighted that when social robots aim at supporting aging-in-place, they benefit from taking advantage of two others fields of robotics: domestic robots and assistive robots.

Moreover, categorizing robots for aging-in-place enable to highlight some challenges and future avenues for Human-Robot Interaction research:

- Improve *social engagement*, either by leveraging robot's appearance to improve anthropomorphism [15] or by increasing social intelligence, for example by improving human-robot personality matching [16] and robot's empathy skills [17].

- Support robot's *adaptability to the user and environment*, by enabling robot's customization to user's abilities, preferences and experiences; and by giving the robot more flexible and autonomous navigation capabilities [18]

- Strengthen *evaluation of robot's assistive outcomes*, by relying on robust methodological design, validated outcome measures and sample characteristics [19]

Lastly, we argue that the "ultimate challenge" shared with all categories of robots is to *enhance adoption* over a long-term, by understanding and improving user acceptance of robots. We believe that better robot functionalities, social abilities and assistive effectiveness are key to encouraging a wide adoption of social robots by older adults [20].

# Socially Assistive Robot in Upper-Limb Rehabilitation: Ethical and Philosophical Considerations following a Feasibility Study

Ronit Feingold Polak, Shelly Levy-Tzedek[*]
Recanati School for Community Health Professions Department of
Physical Therapy
Ben Gurion University of the Negev
Beer-Sheva, Israel

[*]corresponding author: shelly@bgu.ac.il[*]

**ABSTRACT**

Applying robots for rehabilitation raises ethical and philosophical considerations that need to be addressed. We developed a gamified system for stroke rehabilitation. As a first step towards testing it in the clinic, we conducted two feasibility studies with healthy young and older adults, to test their preferences and reactions to the robotic system, and to test whether the physical presence of the robot made a difference in their motivation to continue playing the exercise games. These two experiments raised ethical issues such as safety of autonomous systems and engaging in human-like interaction with an inanimate object. We believe that personalization of the interaction can help achieve a balanced relationship between the human and the robot.

**Key words:** socially assistive robots; human-robot interaction; stroke; motivation; gamification; personalization.

## 1 INTRODUCTION

Stroke is a leading cause for long term disability among adults worldwide [1], with up to 75% of stroke survivors having persistent upper limb (UL) sensorimotor deficits [2,3]. These impairments have a significant effect on the person's ability to be independent in activities of daily living (ADL), participation, and as a result, on quality of life [3,4]. Growing evidence indicates that to maximize recovery of a stroke-affected UL, therapists should apply intensive, repetitive task-specific training [5,6], using everyday tasks that are meaningful and already familiar to the person with stroke [7]. In order for the patient to repeat a certain task many times s/he has to be highly motivated and engaged. Frequently, lack of motivation leads to poor rehabilitation outcomes [8]. In the common practice of clinical rehabilitation, applying a high number of repetitions as part of intensive practice is placing a great challenge on the therapist, due to the limited time available in a therapy session. The difficulty in producing many repetitions of the desired movement is even greater when the rehabilitation program ends and the person has to keep training alone. Therefore, it is imperative to devise feasible, alternative methods for long-term rehabilitation in the home and in the community that promote increased use and improved function in the upper extremity [6]. Robotics are a natural category for this endeavor. Most of the robotic devices for upper-extremity rehabilitation used in research and in the clinical field at present are of two main categories: (a) exoskeletons - which move the patient's arm (b) end-effectors in which only the most distal part of the paretic limb is guided [4]. However, a recent meta-analysis [4], reported that although an improvement in a single movement, like extending the elbow, has been shown, most of these devices did not show improvement in the person's ADL performance. A rather novel type of robot is the socially assistive robot (SAR). SARs are being used in different applications with healthy older adults, e.g., to enhance their exercise motivation, as described by Fasola and Mataric [9], and in assisting individuals with ADL in order to improve quality of life, as described by [10]. The works done by Fasola and Mataric [9] and by Swift-Spong et al. [11] suggest that incorporating SARs into repetitive tasks can increase stroke patients' motivation. However, it is not yet known whether this motivation lasts over a long-term interaction with the SAR, and whether it can lead to a functional improvement post-stroke.

Therefore our overall goal is to develop a closed-loop robotic system for post-stroke UL rehabilitation. We plan to start testing the robotic interaction paradigm we designed in the clinic in the first half of 2018. The system includes a gamified set of exercises, which are similar to the type of exercises a patient would be required to do as part of a rehabilitation program. As a first step towards testing it in the clinic, we conducted two feasibility studies with healthy young and older adults, to test their preferences and perceptions of the robotic system, and to test whether the physical presence of the robot made a difference in terms of their motivation to continue playing the exercise games.

There were several ethical and philosophical issues that participants raised during these experiments, which also lead to ethical considerations that should be taken into account when designing interactions in rehabilitation, with an especially vulnerable population. First, we give a general outline of the methodology in these two experiments, and then we specify the ethical issues that came up, and those that should be considered in the future.

## 2 MATERIALS AND METHODS

60 individuals (30 young and 30 older adults) participated in two experiments. Each experiment included a set of 10-12 games played with a humanoid Robot (Pepper robot, Aldebaran Soft Bank robotics), or with a computer screen. The participants were asked to arrange a set of colored cups according to an image shown on the tablet of the robot, or on a computer screen. Once they completed the task, they were asked to either pet the robot's hand, or touch its tablet. If they played with the computer screen, they indicated the end of a trial by pressing the space key on the computer's key board. At the end of each experiment, the participants were asked to complete a costume-made questionnaire regarding their preferences and perceptions of the interaction. All participants signed an informed consent in accordance with the requirements of the Ben Gurion University Ethics Committee.

## 3 RESULTS

Even though we found a significant preference for interacting with a robot over a screen (p<0.001), participants experienced disappointment from the robot on two levels: first, its slow responses were frustrating, especially for the older population (80% of young adults and 50% of old adults reported they prefer playing with the ROBOT vs. 20% of young adults and 35% of old adults, who preferred the SCREEN condition). The participants indicated that this was an important factor in their motivation to continue interacting with the robot.

Second, when the responses of the robot did not match the participant's perception of his/her performance (e.g., when the robot told the participant that s/he should try to perform the task faster next time – even though the participant performed at his/her highest speed), participants were frustrated – that either they were not good enough, or that the robot's response was not correlated with the performance.

One of the goals of these experiments was to study what type of touch interaction is desired with the robot. While the majority of participants (70% in the first experiment, p=0.12. In the second experiment p=0.008) preferred to touch the hand of the robot, since they reported it felt more "intimate" and "human-like", at least eight participants (13%) reported that it was strange to touch a robot like we would touch a human, and that they did not feel comfortable performing human-like interactions with an inanimate object.

## 4 DISCUSSION & ETHICAL Considerations for Future Experiments

The responses of the participants in these two experiments bring to mind the ethical and philosophical question of what is an appropriate role for a robot in human-robot interactions, and in particular in the context of rehabilitation. We must ask whether it is apt to assign a human-like behavior and appearance to a non-human object, which cannot respond to the users, or the patients, in the way they may expect. This may lead to another source of disappointment and frustration by the users, who may then reject the use of this technology altogether. As Mead & Mataric noted, when interacting with a robot, it is important for the interaction to succeed [12].

We wish to bring to consideration three more ethical issues of the use of rehabilitation robotics. The first is the fear of humans from a de-humanized society, where robots become care givers, and we lose interpersonal relations, as well as the benefits of human touch. As was noted by [13], touch, particularly with another person, is central in building the foundations of social interaction, attachment, and cognition, and early, social touch has unique, beneficial neurophysiological and epigenetic effects.

The second is the question of whether it is medically responsible to have a patient after stroke be in a room with an autonomous robot without human supervision. What if the robot malfunctions? And who is responsible for the outcome of this malfunction. Iosa et al. referred to the safety of therapeutic robots in several aspects: Its medical safety; the need to protect the physical integrity of the patient i.e. by producing the right physiological movement; and other possible harms to the patient like wasting his\her time on an ineffective robot. They concluded that the robot should be at list as safe and effective as other treatments [14].

Lastly, as robots are still an expensive commodity, is it ethical to design a rehabilitation protocol which is only affordable to those with sufficient funds to pay for it?

We believe that these are questions we should consider and discuss in parallel, and not in a sequence, when developing human-robot interactions for therapy. It appears that the answer to at least some of these questions is that there is no single answer, and as was highlighted by Kashi & Levy-Tzedek [15], personalization of the interaction can help achieve a balanced relationship between the human and the robot.

# A Social Robot to Assist in a Therapy for the Care of People with Dementia

Dagoberto Cruz-Sandoval & Jesus Favela
Computer Science Department, CICESE
Ensenada, Mexico
dagoberto@cicese.edu.mx, favela@cicese.edu.mx

## ABSTRACT

People with dementia can benefit from participating in recreational activities. Social robots are increasingly being used to support various kinds of therapies involving people living with dementia. In this work, we present an interactive robot therapy, guided by the robot Eva, based on music and conversation therapy to engage people living with dementia. Furthermore, we suggest dementia care opportunities such as using the robot as a recreational system, dealing with some disruptive symptoms and supporting for caregivers work. The proposed robot therapy has been successfully used by five people with dementia for two months.

## KEYWORDS

Robot therapy, people with dementia; recreational system.

## 1 INTRODUCTION

According to activity theory, older adults benefit from doing as many of the activities that they enjoy as possible as they age [1]. Recreational activities are defined as those that are, to an individual, meaningful, or enjoyable. For people living with dementia, this might include activities such as reminiscence (defined broadly as recalling memories) or listening to music. Activity programs can help people with dementia (PwD) manage symptoms such as agitation, restlessness, and irritability and should be utilized before pharmacological approaches [2].

Robot therapy (RT) has emerged as a non-pharmacological intervention to promote the social contact with PwD. These interventions use robots to simulate human and animal contact [3]. Mainly social assistive robots (SAR) are used in RT. SAR is primarily concerned with robots that provide assistance through social rather than physical interaction [4]. Paro is a robot for pet-therapy applications which has been successfully used in nursing homes. Experimental results suggest that Paro may be effective for reducing stress in nursing-home residents [5]. The robot Brian 2.1 can engage elders in both self-maintenance and cognitively stimulating leisure activities. The robot enacts appropriate assistive behaviors based on the state of the activity and a person's user state [6].

A recreational system uses technologies to support people engaging in recreational activities such as social interactions with friends and family or playing games [2]. We propose the use of SAR as a recreational system to assist an RT for PwD. In this paper, we present an intervention based on robot therapy. Throughout six sessions a SAR promoted an interaction based on music and conversion therapy. Preliminary results suggest that PwD successfully adopted the RT. Moreover, the SAR can be used to deal with problematic behaviors in real-time, and assist caregivers' work on specific tasks.

## 2 Robot therapy-based intervention

We conducted an intervention composed of six sessions per group. The study aimed to explore the interaction between a social robot and PwD, and how it could benefit the PwD. The SAR conducted a group therapy session (often three PwD per group) in a geriatric residence which combined music and conversation therapy elements. We employed a Wizard and Oz model - a human-centric method that uses real technology in a real environment.

During the study, we employed a conversational robot prototype, named Eva, which is a 30 cm anthropomorphic robot [7]. Eva has autonomous features such as natural language understanding, talking, facial expression, and playing music. Furthermore, a human operator can manipulate the behaviors supported by Eva via a web application.

### 2.2 Participants

Caregivers at the geriatric residence proposed the most viable residents to participate in the study. A total of 11 PwD participated in the study. However, 6 of them participated in maximum two sessions - since these residents were often unavailable for different reasons. Thus, we focused on 5 participants who participated in at least five sessions. This group is formed by four women and one man, aged between 71 and 90 (M=82.80, SD=7.19). The scores of Mini-Mental State Examination (MMSE) of them range between 15 and 20 (M=15.40, SD=2.97), which suggest moderate dementia.

### 2.4 Procedure

We divided participants into two groups; group A was composed by P1, P2, P3, while group B was composed by P4 and P5. Following the recommendations to introduce a recreational system in a dementia population, we include a facilitator (a caregiver) on the setup of the session. Moreover, a member of our research team operated the robot – Wizard and Oz model, the operator was out of the spotlight of participants, but he had direct audiovisual feedback.

During the session, the robot Eva enacted an active music therapy, in which participants are actively involved in musical improvisation with instruments or voice. Thus, Eva motivated them to sing, clap, and talk. Moreover, Eva tried to maintain a conversation with participants using conversation strategies to promote conversations with PwD [8]. For example, by using topics that participants enjoy talking about. The most popular topics used by Eva were related to their hometown, music, family, and food.

We conducted a total of 12 sessions – 6 per group. Sessions were conducted weekly, and each session lasted approx. 30 minutes (M=36:08, SD=4:12) and was video-recorded.

## 2.5 Data analysis

We coded the video-recorded sessions during the study to analyze the interaction between participants and the robot Eva. Based on cues for pleasure proposed by the Affect Rating Scale (ARS) [9], we amounted the number of expressions of enjoyment such as smile, laugh, clapping hands, nodding, singing, and dancing. We also counted the number verbal interaction between participants and the robot. Besides, we used an open coding scheme to discover relevant aspects of the interaction.

## 3 RESULTS

Results show an average number of 63.17 utterances from participants per session (Figure 1). While for expressions of enjoyment, the average per session was 19.80 (Figure 2). A positive Pearson correlation value of 0.93 was obtained for direct utterances and a negative correlation of -0.28 for expressions of enjoyment. The number of utterances rose gradually throughout the study. This lead to a reduction of songs used in each session. For group A, the percentage of the session in which music was playing in S1 was 68% (8 songs) while the percentage of time of music for S6 decreased to 17% (2 songs). In group B, the amount of time playing music decreased from 47% (6 songs) to 9% (1 song) in this same period.

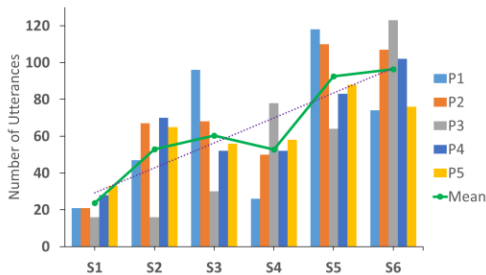

**Figure 1: Direct Utterances from Participants to the Robot.**

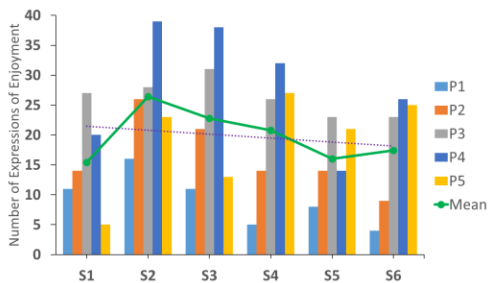

**Figure 2: Number of Expressions of Enjoyment.**

The participants heeded Eva's comments, mostly to participate in the music therapy singing and clapping hands. Furthermore, the robot Eva dealt with unexpected situations, for example when a participant (P4) tried to leave the session - she argued that her family came to visit her which was false, Eva persuaded her to stay in the session:

*P4: I'm leaving because someone came to visit me, but I do not know who is [she stood up and went to the door].*
*Eva: P4 do not leave.*
*P4: Did she say 'goodbye P4'? [to another participant]*
*Eva: Please, stay with us.*
*P4: She is saying 'please stay', how can I refuse that? [P4 sat down].*

Before session 5, P1 experienced anxiety, because she argued that it was time for her to go home. However, during the session, P1 frequently interacted with the robot (118 utterances and 8 enjoyment expressions). She relaxed and calmed while singing and talking about her hometown and food, and she did not speak about going to her home.

## 4 CONCLUSIONS

Preliminary results suggest the adoption by the PwD of the therapy guided by Eva. The therapy was composed of two activities that participants found enjoyable: listening to music and a conversation about topics of interest to them. However, each person with dementia experiences the illness differently. This can produce different ways of interacting with a social robot; it is even possible for a person to refuse to engage with the robot altogether (although this did not happen with any of the 11 participants in our study). Moreover, some PwD may prefer listening to music than conversing, or vice-versa. However, our results suggest that a robot-assisted intervention that combines enjoyable elements for PwD (like music) is an initial step to promote richer interactions. In addition, it is transcendental to define to the social robot as a tool to support caregiver's work, i.e., we consider that the caregivers (familiar) must be responsible for deciding when and how to enact the intervention via the social robot.

# Defining Social Robot Roles in Rehabilitation Based on Observed Patient-Therapist Interactions in Stroke Therapy

Michelle J. Johnson[*]
University of Pennsylvania
Philadelphia, Pennsylvania
Michelle.Johnson2@uphs.upenn.edu

Mayumi Mohan[†]
Max Planck Institute for Intelligent Systems
Stuttgart, Germany
maymohann@is.mpg.de

Rochelle Mendonca[‡]
Temple University
Philadelphia, Pennsylvania
tuf32907@temple.edu

## ABSTRACT
In this paper, we discuss how to improve robot-patient interactions in task-oriented stroke therapy, which currently do not accurately model therapist-patient interactions in the real world. From observations of patient-therapist interactions in task-oriented stroke therapy captured in 8 videos, we describe three dyads of interactions where the therapist and the patient take on a set of acting states or roles and are motivated to move from one role to another when certain physical or verbal stimuli or cues are sensed and received. We propose possible model for robot-patient interaction and discuss challenges to its implementation, including the ethics.

## CCS CONCEPTS
• **Computer systems organization** → **Robotics**; • **Human-centered computing** → *Human computer interaction (HCI)*;

## KEYWORDS
Robotics, Socially Assistive Robots, Stroke Rehabilitation, Human-Human Interaction, Human-Robot Interaction

## 1 INTRODUCTION
By 2030 about 10.8 million older adults will be living with disability due to a stroke. Providing a good quality of life for these older adults requires maximizing independent functioning after a stroke. This implies that in the future, more stroke rehabilitation must occur outside the traditional clinical setting and in more community-based settings such as adult daycare centers, independent living and assisted living centers. Robots can play a unique role in supporting independent living and stroke rehabilitation in non-traditional settings [1, 3–7].

[*]Michelle is with the Departments of Rehabilitation Medicine and Bioengineering
[†]Mayumi is now with the Haptic Intelligence Department
[‡]Rochelle is with the Department of Rehabilitation Sciences

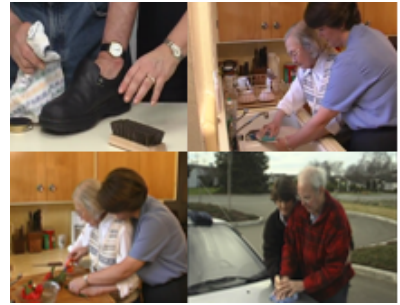

**Figure 1: Occupational therapy videos coded**

Evidence suggests it is appropriate to consider robots as an advanced tool to be used under the therapist's direction, a tool that can implement repetitive and labor-intensive therapies [7]. Current robot-patient interactions do not accurately model therapist-patient interactions in task-oriented stroke therapy, which restricts autonomous robotic interaction. This workshop paper will describe work done to understand patient-therapist interactions, identify roles and key behaviors and propose a model for robot-patient interactions in task-oriented therapy.

## 2 METHODS
We analyzed patient-therapist interactions in task-oriented stroke therapy captured in 8 videos. The videos were independently coded by two therapists who looked for various roles that the therapist and the patient went through during therapy and the cues that causes these role changes using Multimedia Video Task Analysis (MVTA) software. A custom MATLAB script identified the frequency of occurrence of each cue and role. Coder agreement for roles and cues were determines using Cronbach's Alpha ($\alpha$)[2]. The coders examined the videos for the occurrence of three therapist roles (helper, demonstrator and observer) and any physical or verbal cues that may have triggered these roles. For example, in the training of a drinking task, a therapist verbal cue would command or encouragement while a physical could be to reach out to support or lift the impaired arm. In contrast, the patient's verbal cue may be a request for help or clarification while a physical cue would be "does not lift."

## 3 RESULTS

The coders were consistent in identifying roles and cues. Cronbach alpha values for roles and physical and verbal cues were greater than $\alpha$ = 0.95. Both the therapist and the patient spent time in all three roles, 6 times as demonstrator/client observer, 33 times as observer/client performer, and 34 times as helper/client performer with assistance. The frequency and duration of the therapist and patient roles correlated suggesting that treating the dyad as a unit is accurate. Therapists spent more time in the helper role (52.02%), which was especially true when the patient was low functioning. Correspondingly, the patient spent 50.42% of the time being helped. If the patient was able to complete tasks more autonomously (42.81%), then the therapist was in the observer role (41.4%). The therapist performed a total of 190.5(mean between coders) physical cues across all sessions. Most of these cues provided physical assistance to the patient. Out of the 11 physical cues, the reaches, lifts and stabilizes cues were the main ones that caused therapist role changes. The supports/expresses agreement understanding or willingness, requests/asks, commands and states verbal cues had high frequencies of 59.5%, 38.5%, 30.5% and 59.5% respectively.

Based on the results of this study, we overlaid on a stimulus-response (S-R) paradigm. In the S-R model, the therapist and the patient take on a set of acting states or roles and are motivated to move from one role to another when certain physical or verbal stimuli or cues are sensed and received. Figure 2 shows the therapist and the patient as co-actors who go through 3 related sequences of behaviors. For example, in a given therapy session to support relearning of an activity of daily living (ADL), the therapist may begin in the demonstration role. He or she shows the patient how to complete the task, while the patient observes how to do it. The transition from this first dyad to the second one occurs with the "end of task cue" and the when the patient begins the task. In the second dyad, the therapist becomes the observer and the patient the performer. If the patient makes a performance error, coded and observed by the therapist as a physical cues such as "does not reach" and "does not lift", the therapist may transition to the first or third dyad relationship.

## 4 DISCUSSION

We examined how the model varies across 8 activities of daily living tasks and mapped this to a possible stimulus-response paradigm for robot-patient interaction. If a humanoid robot replaces the therapist, then the S-R paradigm dictates that the robot tasks on these three roles as demonstrator, observer and helper and co-acts with the patient. The demonstrator and observer roles are most often found in socially assistive robots (SAR). Coaching SAR systems such as Bandit [6] exhibits these behaviors and is able to provide real-time exercise coaching and rehabilitation to elders and stroke survivors. The helper role is less seen in SARs. The helper role is often seen in hands-on end-effector-based or exoskeleton-based robot therapy systems. Fluid transitioning from contact to non-contact with a patient is not often done due to huge safety concerns about soft and hard impacts. The recent developments in soft robotics makes this type of SAR more attainable.

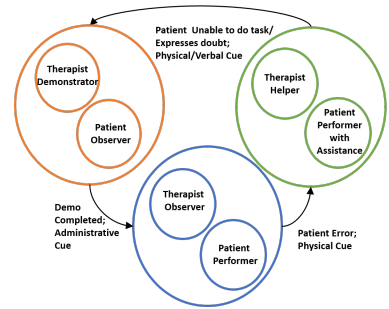

**Figure 2: Three Patient-Therapist Dyadic relationships that frequently appear during task-oriented stroke therapy**

## 5 ETHICAL IMPLICATIONS

Two major ethical implications arise: 1) the appropriateness and effectiveness of the interactions and 2) the implications of replacing patient-therapist interactions with patient-robot interactions. We presume that using actual patient-therapist interactions, across patients of diverse abilities, to design the robot actions will improve interactions between the robot and the patient. If the robot is fully autonomous, will the user acceptance suffer? The second ethical concern should concern the length of time that the robot therapist is used in place of the therapist. A previous study indicated that patients preferred humans to oversee tasks that they perceived as being high-risk, while they were comfortable with a robot performing low risk tasks. Given these ideas, strategy to promote acceptance pf these SARs capable of contact and non-contact therapy would be to ensure shared control of the robot by both the clinician and the patient, which allows them, at all times, to modify or stop encounters that may seem high risk.

## 6 CONCLUSIONS

In conclusion, we intend to use this model to describe how robot-patient interactions in task-oriented therapy could be governed using the roles and cues defined from the stimulus-response model applied to the therapy tasks. In addition, we discuss the challenges and considerations if this model was to be implemented on a humanoid robot.